\documentclass[letterpaper, 10 pt, conference]{ieeeconf}  

\IEEEoverridecommandlockouts                              

\usepackage{graphicx}
\usepackage{tikz}
\usepackage{svg}
\usepackage{url}
\usepackage{amsmath}
\usepackage{amsfonts}
\usepackage{algorithm}
\usepackage[noend]{algpseudocode}

\title{\LARGE \bf 
Learning Which Side to Scan: Multi-View Informed Active Perception with Side Scan Sonar for Autonomous Underwater Vehicles }

\author{Advaith V. Sethuraman$^{1}$, Philip Baldoni$^2$,  Katherine A. Skinner$^1$ and James McMahon$^2$
\thanks{The work of A.V. Sethuraman was supported by the Naval Research Enterprise Intern Program (NREIP). The work of J. McMahon and P. Baldoni is supported by the Office of Naval Research through the NRL Base Program}
\thanks{$^1$A.V. Sethuraman and K. Skinner are with the Department of Robotics, University of Michigan, Ann Arbor, MI 48109 USA  {\tt\footnotesize \{advaiths, kskin\}@umich.edu}.}
\thanks{$^{2}$P. Baldoni and J. McMahon are with the US Naval Research Laboratory, Acoustics Division, Code 7135, Washington D.C., USA
        {\tt\footnotesize \{james.mcmahon, philip.baldoni\}@nrl.navy.mil}.}%
}

\begin{document}

\maketitle
\thispagestyle{empty}
\pagestyle{empty}

\begin{abstract}
Autonomous underwater vehicles often perform surveys that capture multiple views of targets in order to provide more information for human operators or automatic target recognition algorithms. In this work, we address the problem of choosing the most informative views that minimize survey time while maximizing classifier accuracy. We introduce a novel active perception framework for multi-view adaptive surveying and reacquisition using side scan sonar imagery. Our framework addresses this challenge by using a graph formulation for the adaptive survey task. We then use Graph Neural Networks (GNNs) to both classify acquired sonar views and to choose the next best view based on the collected data. We evaluate our method using simulated surveys in a high-fidelity side scan sonar simulator. Our results demonstrate that our approach is able to surpass the state-of-the-art in classification accuracy and survey efficiency. This framework is a promising approach for more efficient autonomous missions involving side scan sonar, such as underwater exploration, marine archaeology, and environmental monitoring.

\end{abstract}

\section{Introduction}


Autonomous underwater vehicles (AUVs) are widely used for search, detection, and recognition of underwater objects. AUVs perform these tasks using a variety of acoustic sensors including side scan sonar \cite{sun2021review}. Side scan sonars are commonly characterized by their high along-track resolution, which refers to the level of detail that can be seen in the image, as well as the wide swath width, which refers to the width of the area that is imaged \cite{medwin1999,blondel2010}. 
\begin{figure}
\includegraphics[width=\linewidth]{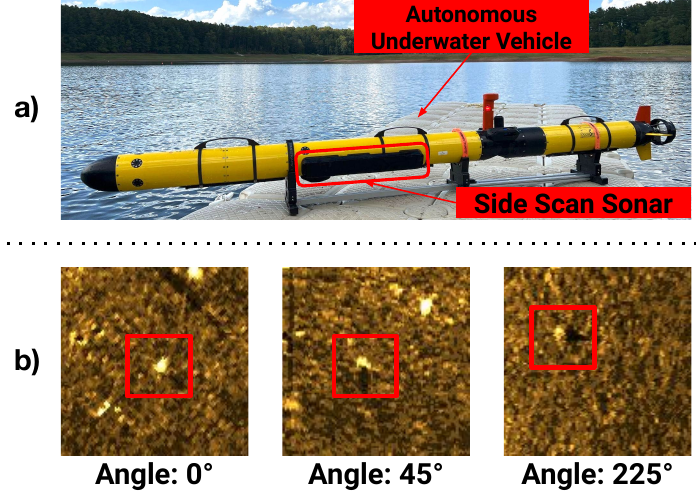}
\caption{a) An Iver3 autonomous underwater vehicle equipped with a Klein 3500 side scan sonar system. b) A real cylindrical target imaged in side scan sonar is shown in the red boxes. Side scan sonar image appearance is highly dependent on viewing angle and target geometry, both of which determine how much acoustic energy returns to the receiver. \label{teaser}}
\vspace{-6mm}
\end{figure}
There are three important physical properties of side scan image formation that pose challenges to automated classification of objects underwater. First, the backscatter from sound interacting with an object can be approximated as Lambertian \cite{Urick1956,mckinney1964}. This means that depending on the geometry of the object and viewing angle of the sensor, sound may scatter away from the receiver array and never be recorded. Next, similar to RGB-D sensors or LIDAR, acoustic shadowing occurs when objects and terrain obstruct the propagation of sound underwater. Depending on the terrain type and clutter around an object, these shadows could help or hurt classifier performance. Finally, side scan sonar accumulates all the returns from a given range into the same pixel bin on the image. As a result, there is a natural elevation angle ambiguity that can cause objects to appear differently based on the viewing angle \cite{woock2010}. 

Due to the view-dependence of side scan imagery, when trying to classify an object using side scan sonar, an AUV operator will typically program a vehicle trajectory that exhaustively inspects the object from multiple viewing angles. Then, Automatic Target Recognition (ATR) algorithms can classify the object based on a variety of hand-crafted features or features learned through machine learning. This process is called \textit{reacquisition} and can require in excess of 30 minutes per object depending on survey parameters. An example of the view dependence of side scan imagery is shown in Fig. \ref{teaser}. 

To address the inefficiency of exhaustive reacquisition, this work aims to develop a novel method to find and navigate to the most informative views for an ATR algorithm to produce an accurate classification. We call this task \textit{Adaptive Surveying and Reacquisition} (ASR).
This work presents a graph formulation for both the multi-view classification problem and the next best view problem that allows us to solve the ASR task more quickly and with higher accuracy than state-of-the-art adaptive survey planning algorithms \cite{imvp}. 

The main contributions of this work are: 

\begin{itemize}
    \item We present a novel perception framework called the \textit{angular view-graph} for multi-view adaptive surveying and reacquisition in side scan sonar using graph neural networks. 
    \item We present a novel reward function that encourages a reinforcement learning agent to choose the minimum number of next best views necessary for an accurate classification. 
    \item We present an ASR algorithm for AUVs that finds and classifies all targets in a search region without pre-survey knowledge of target locations. 
    \item We report extensive results from a photorealistic sonar simulation environment that demonstrate that our method outperforms state-of-the-art adaptive survey methods in classification accuracy, classification efficiency, and coverage rate. 
\end{itemize}
Our experiments demonstrate that learning to choose the next best view can significantly expedite underwater search and classification tasks, many of which operate under limited time budgets.


\begin{figure*}
\centering
\vspace{7mm}
\includegraphics[width=0.95\linewidth]{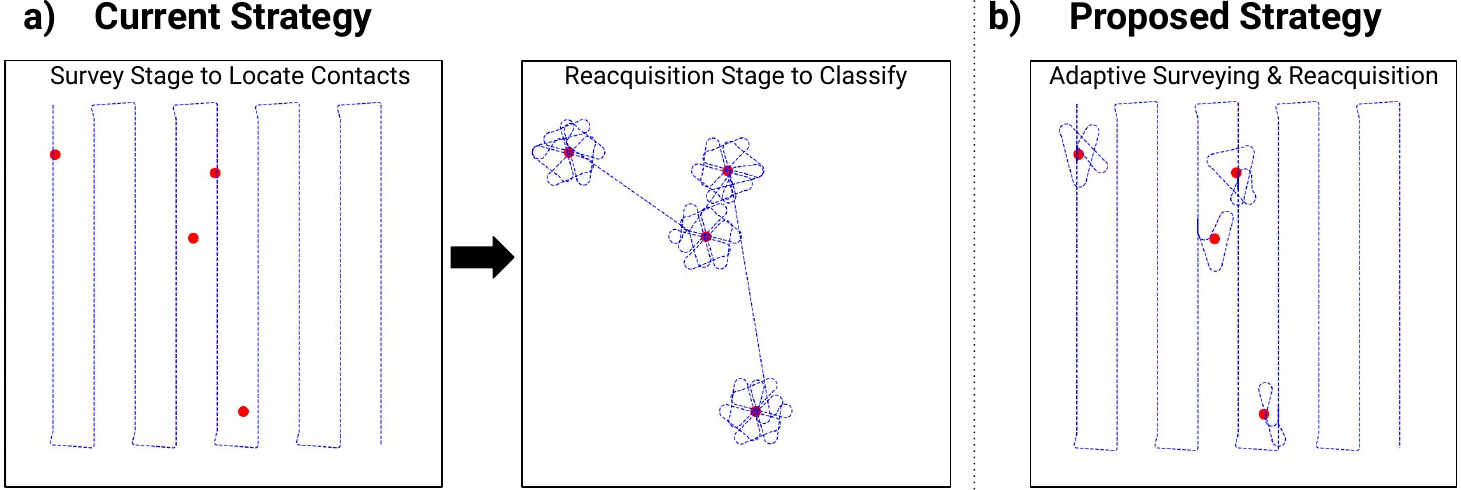}
\caption{a) The traditional Surveying and Reacquisition process. First a comprehensive survey of the region is performed (lawnmower pattern shown), then the discovered contacts (shown as red dots) are reacquired and inspected. b) The proposed strategy in this paper is to combine the Surveying and Reacquisition stages and inform the Reacquisition planner with the most informative views. By avoiding uninformative reacquisition survey legs, we can reduce the total time for clearing a search region. \label{lm}}
\vspace{-4mm}
\end{figure*}
\section{Related Work}
\subsection{Graph Neural Networks for Robotics}
Recent work in robotics has leveraged the inductive biases of graph neural networks (GNNs) to handle variable sized inputs and aggregate information across nodes effectively \cite{yixuan-icra2023-graph-relations, kumar_gnn, gnn_hron}. \cite{yixuan-icra2023-graph-relations} uses GNNs to better reason about the interactions between objects in multi-object manipulation tasks. \cite{gnn_hron} explores the use of heterogeneous edges in scene-graphs to describe the relations between objects in a scene and attention mechanisms to expedite embodied search for objects. In multi-agent coordination, \cite{kumar_gnn} uses a graph-based imitation learning policy for perimeter defense that exploits each agent's local measurements and communications with other agents. 
In this work, we propose a graph formulation for the ASR problem. Our \textit{angular view-graph} encodes information from sonar views as nodes and the angular relation between views as edges. To the best of our knowledge, we are the first to propose this graph formulation for both multi-view classification and next best view planning. 

\subsection{Multi-View Classification}
Multi-view classification is a computer vision task that considers multiple views of an object to produce a final classification \cite{su15mvcnn, view-gcn, rotnet, view-gcn2}. Prior methods embed each view of an object into a feature space, then apply an aggregation function to produce a final classification. 

Multi-View Convolutional Neural Networks (MVCNNs) take 12 views of an object and apply a max-pooling operator to the embedded features to produce a final classification \cite{su15mvcnn}. Although this approach effectively combines data across multiple views, it cannot handle partial views nor does it provide information about the relations between each view. Although follow-up work such as RotationNet \cite{rotnet} addresses the partial view problem and regresses the viewing parameters, it is less effective in classification than recent work that uses graph formulations. 

View-GCN explores the use of view-graphs for multi-view classification \cite{view-gcn, view-gcn2}. 
Since View-GCN's view-graph only connects the k-nearest neighbor views, we find that its formulation is not immediately applicable to our robotics task. In robotic surveys, the k-nearest neighbors of a view may not be captured yet, yielding an uninformative view-graph with minimal edge connectivity. 

In contrast to the view-graph from View-GCN, we propose an \textit{angular view-graph} formulation, which is a fully connected graph in which the edges are the angular offsets between each view. This formulation allows us to exploit the benefits of the original view-graph representation while also being descriptive enough in partial view scenarios. We find experimentally that our angular view-graph is a more effective structure for the ASR task. 

\subsection{Next Best View Planning}
The next best view planning problem is a sub-problem in active perception that is concerned with finding view parameters that achieve a desired goal \cite{Lauri2020MultiSensorNP, menon2023nbvsc, 9811800, peralta2020next, 9340916}. Lauri et al. use a ray-tracing-based view utility function and a greedy algorithm to select the next best view for scene reconstruction with multiple robots \cite{Lauri2020MultiSensorNP}. NBV-SC avoids ray-tracing and instead leverages shape completion prediction to determine the next best view for reconstructing fruits \cite{menon2023nbvsc}. Alternatively, \cite{9811800, peralta2020next} use reinforcement learning to find the next best view with reward functions that encourage coverage of unseen areas using 3D sensors. While most prior work focuses on maximizing 3D coverage, in this work we aim to maximize classification accuracy while minimizing survey time, which presents a fundamentally different task. 


It has been found that actively positioning sensors to capture alternative views can improve classification performance \cite{le2008active, johns2016pairwise}. \cite{le2008active} uses a Gaussian Process Regression and a utility function based on Mutual Information to choose the next viewing angle for an RGB camera. More recently, \cite{johns2016pairwise} considers the problem of choosing the next best trajectory of views for a camera that increases classification accuracy. However, this approach is not immediately applicable to our problem because one side scan image corresponds to a single pass and the trajectory between passes does not produce useful re-observations of the target. 

Adaptive view planning for acoustic sensors has been explored in the past \cite{imvp, 5664609}. Myers and Williams use a POMDP formulation to find the best views for classification in Synthetic Aperture Sonar (SAS) imagery \cite{5664609}. However, their work focuses on improving classifier accuracy rather than multi-target survey time and their reward function does not consider the cost of capturing additional views. IMVP is a framework that uses a Bayes Net to infer and navigate to the most informative views for side scan sonar \cite{imvp}. However, the IMVP formulation uses hand chosen features that are modeled as categorical random variables. This choice restricts the expressiveness of the system, yielding poor performance when features are not discriminative enough between challenging classes. Instead, our method uses a graph neural network to learn an aggregation function for classification. We then choose the next best view in a data-driven manner using deep reinforcement learning, which exploits the full expressiveness of the proposed angular view-graph representation. 


\section{Technical Approach}
\begin{figure*}
\centering
\includegraphics[width=\linewidth]{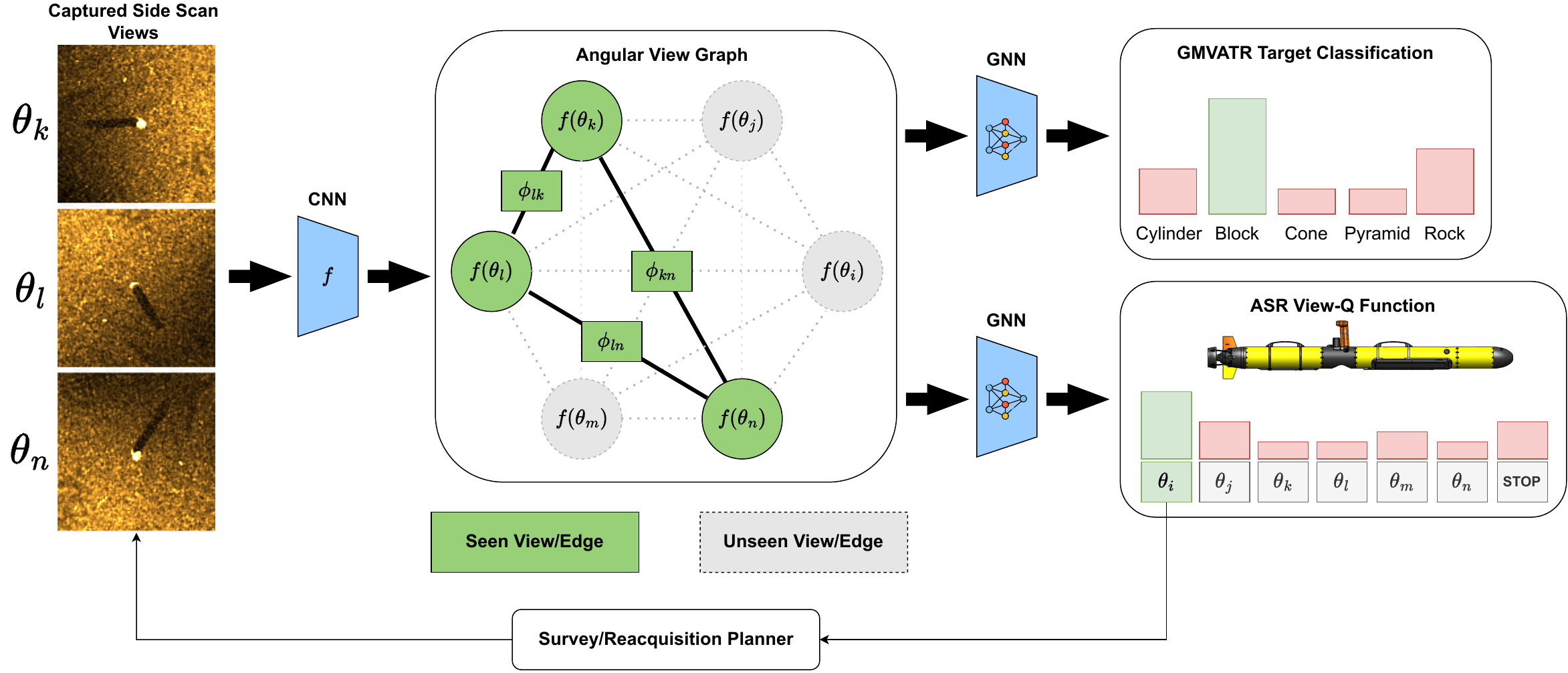}
\caption{Our proposed framework. For a side scan view captured at an angle of $\theta$ degrees, we produce feature embedding $f(\theta)$ using a CNN. Then we form the angular view-graph which consists of the feature embeddings and angular constraint edges $\phi$.  The graph is used for both Multi-View Automatic Target Recognition tasks and as the state of a reinforcement learning (DQN) agent. The DQN agent informs the planner of the next best views to capture. The process is repeated and the graph is incrementally expanded until the DQN agent stops reacquisition. The target classification decision is computed using the final angular view-graph. \label{arch}}
\end{figure*}

Our proposed framework is shown in Fig. \ref{arch}. First, we compose an angular view-graph (AVG) with captured side scan views. Then, we pass the AVG through a GNN for classification and a GNN for our reinforcement learning policy. Finally, we navigate to and capture the next best view as predicted by our next best view policy. 

\subsection{Problem Formulation}
A common practice for classifying multiple targets within a search region is to first perform a comprehensive survey to find all contacts, then perform reacquisition patterns for close-up inspection and classification as shown in Fig. \ref{lm}a. In this work, we are interested in reducing the total time to clear a search region, which involves 1) finding all contacts within the region and 2) classifying all the contacts that have been found. We call this task Surveying and Reacquisition (SR), and we provide further details of each sub-task below. 

\subsubsection{Surveying}
In the surveying stage, we assume the locations and number of targets $N$ are not known beforehand. It is unlikely that the user will have prior surveys of the region when deploying an AUV in a novel environment. The survey environment consists of $N$ targets placed in an $H \times W$ rectangular search region with no obstacles. In this work, we will require 100\% coverage of the search region, with the simplest coverage plan being the \textit{lawnmower pattern} (LM). The parameters of the lawnmower pattern can be set based on the sonar's max range and the AUV's altitude. 

\subsubsection{Reacquisition} 
After an initial survey is completed, the locations of all contacts are known. The next goal is to capture additional views of the targets for classification. We consider $K$ discrete viewing angles where each is parameterized by an angle $\theta$, pass length $L$, and a radius $R$ to the target location. A straight line pass is required to image the target, as roll/yaw changes create distortion in the sonar imagery. This is called an Object Identification (OID) survey.

\subsubsection{Adaptive Survey and Reaquisition}
In an effort to reduce the time for SR, we consider the \textit{Adaptive Survey and Reaquisition} task, where the AUV can choose the most informative set of $\theta$ for its perception goal, and keep the parameters $L$ and $R$ fixed. An example ASR trajectory is shown in Fig. \ref{lm}b.  

The perception task of interest in this work is multi-class classification of targets in side scan sonar images. The objective of an efficient ASR algorithm is to find and accurately classify all targets in the search region as quickly as possible.

\subsection{Multi-View ATR as an Angular View-Graph}
\subsubsection{Angular View Graph}
We present a novel representation for the ASR task that exploits the inductive biases of GNNs. Consider sonar images for a given target captured at different viewing angles as $I(\theta_i)$, where $\theta_i$ denotes the angle offset with respect to the first contact. An AVG $G$ on $k$ views represents the relationship between $k$ embedded sonar images $[f(I(\theta_1)), \hdots, f(I(\theta_k))]$, where $f(\cdot)$ represents a feature embedding produced by a CNN. $G = (V,E)$ is formally defined as a graph  with vertices $V = \{f(\theta_i), \ i \in [1, \hdots, k]\}$ and adjacency matrix $E \in \mathbb{R}^{i \times i}$ with $E_{ij} = \phi(\theta_i, \theta_j)$ where  
\begin{equation}
\phi(\theta_i, \theta_j) = MLP([\angle e^{i(\theta_j - \theta_i)}, i, j])
\end{equation}
For brevity, we use phasor notation to handle angular wraparound consistently. $MLP: \mathbb{R}^3 \rightarrow \mathbb{R}^+$ is a multi-layer perceptron that consists of two fully-connected layers, with ReLU nonlinearities. A final Sigmoid is used since the edge weights must be non-negative. Note that our proposed AVG is a fully connected graph as opposed to the kNN formulation of View-GCN \cite{view-gcn}. This allows partial views to still be connected and to inform next best view planning and classification. Only captured views are added to the angular view-graph. 

\subsubsection{Multi-View ATR}
To produce classification probabilities from the angular view-graph, we introduce the Graph Multi-View ATR Network (GMVATR). Our GMVATR is composed of a CNN backbone to produce image features and a GNN composed of three GCNConv layers with hidden dimension 64 \cite{kipf2017semi}. Finally, a fully-connected layer produces logits for classification. 

\subsection{View-Q Function}
\subsubsection{Reinforcement Learning Details}
In order to perform efficient surveys, we wish to navigate between the minimum number of views that are informative enough to correctly classify a target. We consider the angular view-graph to be the state of a Markov Decision Process (MDP). An MDP is parametrized by a 5-tuple $(\mathcal{S}, \mathcal{A}, \mathcal{P}, r, \gamma)$.\\
We consider $\mathcal{S} = \cup_i^{K} \ \mathcal{G}_i$, where $\mathcal{G}_i = \{(V, E): |V| = i, E \in \mathbb{R}^{i \times i}\}$ is the set of all fully connected angular view-graphs with $i$ vertices. The discrete action space $\mathcal{A} = \{\frac{2\pi}{K}i, i \in [0,K)\} \cup \{STOP\}$ denotes the set of all viewing angle offsets for a particular target. Note that we define the first contact the robot receives as $0^\circ$. The $STOP$ action allows the agent to stop collecting data at the current target and travel to the next target. We assume a deterministic transition function. 

The goal of the View-Q Function is to find the smallest set of views that produce an accurate classification from the GMVATR algorithm. 
We formulate our reward function at timestep $k$ as: 
$$
r_k =
\begin{cases}
+10 \ \text{if} \ \hat{y} = y\\
-10 \ \text{if} \ \hat{y} \neq y\\
-1 \hspace{3mm} \text{if} \ k > k_{FCC} \\ 
0 \ \text{otherwise}
\end{cases}
$$

\noindent where $\hat{y}$ is the classifier prediction, $y$ is the ground truth label for the target, and $k_{FCC}$ is the timestep of the first correct classification (FCC). Note that we encourage succinct reacquisitions by penalizing every view beyond the FCC with a reward of $-1$. 
\subsubsection{Network Details}

We train a Deep Q-Network (DQN) policy with experience replay in PyTorch. The target and policy networks for the DQN are separate but follow the same architecture. They are composed of one GCNConv layer then two fully-connected layers. The classifier is frozen during the training of the DQN algorithm. We train our RL policy for approximately 3.5 million environment steps. 
\subsection{Adaptive Survey and Reacquisition Algorithm}

To avoid redundant surveying, we propose an ASR algorithm that combines the surveying and reacquisition stages in Algorithm (\ref{MVAS}). Consider $G(V)$ to be a function that creates an angular view-graph for a set of captured views $V$. 

\begin{algorithm}[h]
\caption{Adaptive Survey and Reacquisition Planner\label{MVAS}}
\label{alg:survey_trajectory}
\begin{algorithmic}[1]
\State $LM \gets \texttt{plan\_lawnmower\_traj()}$
\While{\textit{not done}}
    \State go to next waypoint in $LM$
    \If{\textit{contact in ensonified sonar region}}
        \State $\theta_c \gets$ \texttt{get\_closest\_OID\_leg()}
        \State robot captures $I(\theta_c)$
        \State $V \gets \{I(\theta_c)\}$
        \State $a_{NBV} \gets \texttt{V-QF}(G(V))$
        \While{$a_{NBV} \neq STOP$}
            \State robot captures $I(a_{NBV})$
            \State $V \gets V \cup I(a_{NBV})$
            \State $a_{NBV} \gets \texttt{V-QF}(G(V))$
        \EndWhile
    \EndIf
\EndWhile
\end{algorithmic}
\end{algorithm}

\newpage
\section{Experiments and Results}

Through our experiments, we wish to gain insight into the following research questions: 
\begin{itemize}
\item \textbf{Q1:} Does the angular view-graph representation enable better classification performance at a lower cost for the ASR task compared to state-of-the-art multi-view survey planning algorithms? 
\item \textbf{Q2:} How important is choosing the next \textit{best} view versus choosing an arbitrary next view? 
\end{itemize}

\subsection{Side Scan Sonar Simulator}
\begin{figure}
\vspace{2mm}
\centering
\includegraphics[width=0.96\linewidth]{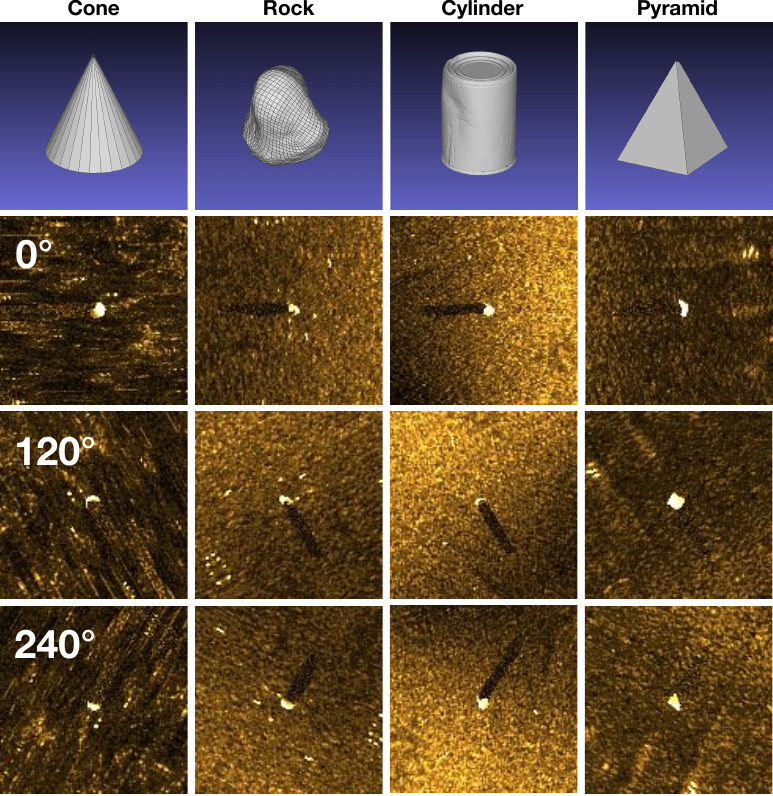}
\caption{Photorealistic side scan sonar images produced by our sonar simulator. Each target's orientation, size, and terrain is randomized, then imaged using a simulated OID pattern with $K=6$, pass length $L=75m$, radius $R=12m$, height from bottom (HFB) of 3m and resolution of 0.02m/pixel. Note how the sonar images change as a function of viewing angle. Side scan sonar images are single channel images of acoustic intensities, but a gold color palette is applied for better visibility. \label{targets}} \vspace{-4mm}
\end{figure}

Since side scan sonar uses a relatively high frequency acoustic source, we are able to simulate sonar image formation using ray-tracing techniques similar to optical computer graphics \cite{bell1995model}. Specifically, we create a side scan sonar simulator written as a CUDA kernel in the Nvidia Optix framework \cite{optix}. Rays are traced from a virtual moving source (AUV) and a custom shader renders the acoustic image using normal information and material specific acoustic reflectance information. Finally, speckle noise is added and the rendered image is composited onto real terrain images. The final resolution of the sonar images is $120 \times 120$ pixels. This corresponds to a $3m \times 3m$ crop region in space. We are able to simulate randomized scenes imaged using OID patterns. Terrain elevation and shadowing is randomized using Perlin noise as a heightmap. Examples of generated sonar images from our simulator are shown in Fig. \ref{targets}. 

In this work, we consider 5 classes of targets: $\{cylinder,\ cone,\ block,\ pyramid,\ rock\}$. For the rock class, we use the Blender Rock Generator Add-On to procedurally generate rocks of similar dimensions to the targets \cite{blender_rock}. Our dataset consists of 8,000 distinct targets each with 6 views for a total of 48,000 side scan sonar images. 

\subsection{Baselines} 
We compare our method to recent informed multi-view path planning algorithms and multi-view classification methods. For baselines without next best view policies, we use random choice of actions (RAND1) and exhaustive (EXH). EXH views all $K$ views. RAND2 is a next best view policy that chooses the same number of views for each target as the V-QF policy but chooses the views randomly. 
\begin{table*}[]
\vspace{4mm}
\centering
\caption{Simulation results ($M \pm SD$) for our method compared with baselines. All metrics are averaged across 3 random seeds with 100 survey simulations each. $(H,W) = (1.2km,\ 1.2km)$ and $N=16$. Best performance is \textbf{\textcolor{blue}{bold blue}} and second best is in \textcolor{red}{red}. \label{sim_res}}
\scalebox{0.92}{
\begin{tabular}{lccccccc}
\hline
\multicolumn{1}{c}{Method} & Accuracy  $\uparrow$           & Time (hr)    $\downarrow$                     & Recall $\uparrow$              & CR ($\text{km}^{2}\text{hr}^{-1}$) $\uparrow$ & CE (hr$^{-1}$) $\uparrow$ & AV $\downarrow$   \\ \hline
Single Pass LM             & $0.628 \pm 0.136$          & \textcolor{blue}{\boldmath{$3.871 \pm 0.016$}} &  $0.558 \pm 0.179$         & \boldmath{\textcolor{blue}{$0.442 \pm 0.002$}}                 &$0.171 \pm 0.055$  & \boldmath{\textcolor{blue}{$1.00 \pm 0.000$ }}         \\
SV + OR + EXH              & $0.852 \pm 0.089$          & $7.380 \pm 0.028$            & $0.802 \pm 0.141$                          & $0.195 \pm 0.001$   & $0.109 \pm 0.023$                   & $6.000 \pm 0.000$  \\
SV + MEAN + EXH            & $0.919 \pm 0.060$         & $7.380 \pm 0.028$                 & $0.890 \pm 0.114 $        & $0.195 \pm 0.001$                         & $0.121 \pm 0.019$    & $6.000 \pm 0.000$                    \\
SV + OR + RAND             & $0.786 \pm 0.112$          & $6.001 \pm 0.219$                   & $0.732 \pm 0.169$          & $0.240 \pm 0.009$      & $0.122 \pm 0.028$                       & $3.620 \pm 1.719$                    \\
SV + MEAN + RAND           & $0.838 \pm 0.087$         & $6.001 \pm 0.219$                &$0.769 \pm 0.156$          & $0.240 \pm 0.009$                         & $0.128 \pm 0.033$   & $3.620 \pm 1.719$                     \\
IMVP \cite{imvp}                      & $0.909 \pm 0.070$          & $5.963 \pm 0.125$                & $0.872 \pm 0.121$          & $0.242 \pm 0.312$   & $0.146 \pm 0.027$                        & $3.576 \pm 1.018$                    \\
View-GCN \cite{view-gcn}          & \textcolor{red}{$0.931 \pm 0.061$}          & $4.880 \pm 0.165$                & \textcolor{red}{$0.910 \pm 0.120$}         & $0.295 \pm 0.015$   & \textcolor{red}{$0.186 \pm 0.026$}  & $1.210 \pm 0.560$                \\ \hline
GMVATR + AVG + V-QF (ours)  &\boldmath{\textcolor{blue}{$ 0.952 \pm 0.066 $}}        & \textcolor{red}{$4.858 \pm 0.172$}               & \boldmath{\textcolor{blue}{$0.930 \pm 0.108$}}          & \textcolor{red}{$0.296 \pm 0.010$}                          & \boldmath{\textcolor{blue}{$0.191 \pm 0.032$ }}             & \textcolor{red}{$1.180 \pm 0.686$} \\ \hline
\end{tabular}}
\end{table*}
\begin{itemize}
\item \textbf{SV\_LM} follows a fixed lawnmower pattern (LM) path and must classify the target using the arbitrary single view captured along the survey path. 
\item \textbf{SV\_OR} uses a single-view (SV) classifier and combines predictions using the rule $p(X) = max_i \ p(x_i)$. 
\item \textbf{SV\_MEAN} uses a single-view (SV) classifier and combines predictions using the rule $p(X) = \frac{1}{K} \sum_{i=1}^{K} p(x_i)$. 
\item \textbf{View-GCN \cite{view-gcn}} uses the same GNN as our method for classification but uses the k-nearest neighbor view-graph formulation from \cite{view-gcn, view-gcn2}. 
\item \textbf{IMVP \cite{imvp}} combines predictions from multiple views using a Bayes update \cite{imvp}. The next best view is chosen by inference on a Bayes Net. 
\end{itemize}

\subsection{Ablations}
We ablate the AVG and replace the View-Q function (V-QF) with a RAND or EXH policy to isolate their respective contributions to overall survey performance. 

\begin{itemize}
\item \textbf{GMVATR} This model uses a fully connected view-graph without any angular information in the edges. 
\item \textbf{GMVATR + AVG} This model (using V-QF) is the final method presented in this paper. 
\end{itemize}
\subsection{Implementation Details}
All networks use a ResNet-50 backbone and are trained from scratch initialized with ImageNet pre-trained weights. We use two augmentation strategies: 1) RandAug and 2) subgraph augmentation. Although RandAug is used primarily for optical imagery, we observed that it still improved network convergence and generalizability. Next, we note that an AVG with $K$ views has $\sum_{i=1}^{K} {K \choose i}$ unique subgraphs. Methodically creating and training on all of these subgraphs also significantly improved generalizability. The loss for each set of subgraphs with $i$ views was weighted inversely by ${K \choose i}$. 
\subsubsection{IMVP}
We re-implement IMVP using the Python library \texttt{bnlearn} to learn the Bayes Net and conditional probability tables from our dataset. We use a classifier with two branches consisting of linear layers that output 5 classes and 4 classes to predict categorical features $\hat{X_1} = \text{SHAPE}, \hat{X_2} = \text{VOLUME}, \hat{Y} = \text{CLASS}$. Finally, we use $\epsilon_{CL}=0.85$ for survey simulations. 
\subsubsection{Path Planning}
Dubins paths with turning radius of $15$m were used for all trajectories. An average speed of $2.0$ m/s is assumed for the simulation. The AUV follows a lawnmower pattern with 0\% overlap using height from bottom (HFB) of $10$m, horizontal spacing of $100$m and a sensor range of $50$m. For reacquisitions, the AUV must dive to a lower HFB of $3$m. 
\subsubsection{Simulator Details}The simulated OID patterns use $K=6$, pass length $L=75$m, radius $R=12$m, and $HFB=3$m. Following \cite{imvp}, each survey has $N=16$ randomly chosen target locations within a search region of $(1.2\text{km},\  1.2\text{km})$.

 
\subsection{\textbf{Q1:} Adaptive Survey and Reacquisition}
We study the performance of our perception and planning architecture in the context of autonomous surveys. We randomly split our sonar dataset into 70/10/20 train/val/test. The reported evaluation metrics are averaged across three random seeds with 100 surveys each.  
\subsubsection{Evaluation Metrics}
We consider metrics commonly used for evaluating autonomous underwater surveys. 

\begin{itemize}
\item \textbf{Accuracy} 
\item \textbf{Total Time (T) [hours]}
\item \textbf{Per-Target Time (PTT) [\text{hours}]:} $\frac{T}{N}$
\item \textbf{Classification Efficiency (CE) [$\text{hours}^{-1}$]:} $\frac{TP}{T(TP+FN)}$
\item \textbf{Coverage Rate (CR) [$\text{km}^2 \ \text{hours}^{-1}$]:} $\frac{HW}{T}$
\item \textbf{Average Number of Views (AV)} 
\end{itemize}

The results of our surveys are shown Table (\ref{sim_res}). Our method outperforms baselines, exhibiting improved accuracy, coverage rate (CR), and classification efficiency (CE). IMVP \cite{imvp} achieves $0.909$ accuracy but at the cost of more than double the average number of views (AV) as our method. This results in much longer survey time and lower CE. Although the Single Pass LM has the fastest survey time, and lowest AV, it performs badly in accuracy/recall since it does not conduct reacquisition and receives only one view of the object. These results illustrate the importance of the reacquisition stage for classification. 

\subsection{\textbf{Q2:} Next Best View Policy}
To understand how important it is to choose the next \textit{best} view, we introduce a new policy \textbf{RAND2}. For a given target, RAND2 chooses the same number of views as our V-QF policy, but chooses the views randomly. 

GMVATR + AVG + RAND2 has reduced accuracy/recall performance compared to our model at the same number of average views. This indicates that it is crucial to choose the next \textit{best} view, and choosing views arbitrarily risks reducing classifier performance. Without the angular view-graph representation, GMVATR + RAND2 achieves significantly worse performance compared to GMVATR + AVG + RAND2, indicating that the addition of the angular view-graph can improve classification performance independently of the optimal next best view policy. Finally, the performance gap between GMVATR + V-QF and GMVATR + AVG + V-QF illustrates the benefit of using our AVG formulation. 
\begin{table}[t]
\centering
\caption{Ablations follow simulation parameters from Table (\ref{sim_res}).}
\scalebox{0.9}{
\begin{tabular}{lcccc}
\hline
Methods              & Accuracy $\uparrow$                                                                           & Recall $\uparrow$                   & AV  $\downarrow$                                         \\ \hline
GMVATR + RAND1       & 0.94                                                                       & 0.92                                                            & 4.59                                         \\
GMVATR + RAND2       & 0.85                                                             & 0.81                                                                      & 1.18                                         \\
GMVATR + EXH         & 0.97                                                                    & 0.97                                                                        & 6.00                                         \\
GMVATR + V-QF        & 0.92 & 0.89 & 2.35  & \\
GMVATR + AVG + RAND1 & 0.97                                                                       & 0.94                                                              & 4.59                                         \\
GMVATR + AVG + RAND2 & 0.92                                                                        & 0.90                                                   & 1.18                                         \\
GMVATR + AVG + EXH   & 0.99                                                               & 0.98                                                  & 6.00                                         \\ \hline
GMVATR + AVG + V-QF    & 0.95  & 0.93    & 1.18                                         \\ \hline
\end{tabular}}
\vspace{-4mm}
\end{table}

\section{Conclusion}
We present a novel active perception method for side scan sonar that 1) chooses the next best view to maximize classifier performance and 2) aggregates and classifies multiple views from an adaptive survey. Through extensive experiments, we show that our method achieves superior accuracy, coverage rate, and classification efficiency compared to state-of-the-art methods. We also studied the impact of the proposed angular view-graph representation and showed that it is a useful structure for the next best view task of interest. 

Future work will consider sim2real transfer \cite{hofer2021sim2real,Sethuraman_2023_BMVC} to facilitate real-world trials. The effect of high-clutter environments and acoustic phenomena (e.g. artifacts, distortions, multi-path) on GMVATR and V-QF requires further validation. Finally, expanding the action space of V-QF to include pass length $L$ and radius $R$ may allow more flexible, efficient, and emergent reacquisition trajectories.





\section{Acknowledgements}


We would like to acknowledge Bill Dryer Sr. and Bill Dryer Jr. for facilitating field work experiments. We would also like to acknowledge Jim Turner for insightful discussions regarding multi-view classification, and David Raudales for providing practical advice about ATR surveying. 
\newpage
\bibliographystyle{ieeetr}
\bibliography{ref}
\end{document}